\pdfoutput=1

\documentclass[11pt]{article}
\usepackage{acl2014}
\usepackage{times}
\usepackage{graphicx}
\usepackage{latexsym}
\usepackage{amsmath}
\usepackage{amssymb}
\usepackage{paralist}
\usepackage{booktabs}
\usepackage{tikz}
\usepackage{multirow}
\usepackage{url}
\usetikzlibrary{matrix}

\renewcommand{\vec}{\mathbf}

\title{A Convolutional Neural Network for Modelling Sentences}

\author{Nal Kalchbrenner \\\And Edward Grefenstette\\ $ $ \\ {\tt\small \{nal.kalchbrenner, edward.grefenstette, phil.blunsom\}@cs.ox.ac.uk}\\{Department of Computer Science} \\{University of Oxford}\\\And
  Phil Blunsom \\
\\}

\date{}

\begin{document}
\maketitle
\begin{abstract}

The ability to accurately represent sentences is central to language understanding. We describe a convolutional architecture dubbed the Dynamic Convolutional Neural Network (DCNN) that we adopt for the semantic modelling of sentences. The network uses Dynamic $k$-Max Pooling, a global pooling operation over linear sequences. The network handles input sentences of varying length and induces a feature graph over the sentence that  is capable of explicitly capturing short and long-range relations. The network does not rely on a parse tree and is easily applicable to any language. We test the DCNN in four experiments: small scale binary and multi-class sentiment prediction, six-way question classification and Twitter sentiment prediction by distant supervision. The network achieves excellent performance in the first three tasks and a greater than $25\%$ error reduction in the last task with respect to the strongest baseline.

\end{abstract}

\section{Introduction}

The aim of a sentence model is to analyse and represent the semantic content of a sentence for purposes of classification or generation. The sentence modelling problem is at the core of many tasks involving a degree of natural language comprehension. These tasks include sentiment analysis, paraphrase detection, entailment recognition, summarisation, discourse analysis, machine translation, grounded language learning and image retrieval.
Since individual sentences are rarely observed or not observed at all, one must represent a sentence in terms of features that depend on the words and short $n$-grams in the sentence that are frequently observed. The core of a sentence model  involves a {feature function} that defines the process by which the features of the sentence are extracted from the features of the words or $n$-grams.

\begin{figure}
\centering
\label{DAGs}

{\includegraphics[width=0.48\textwidth]{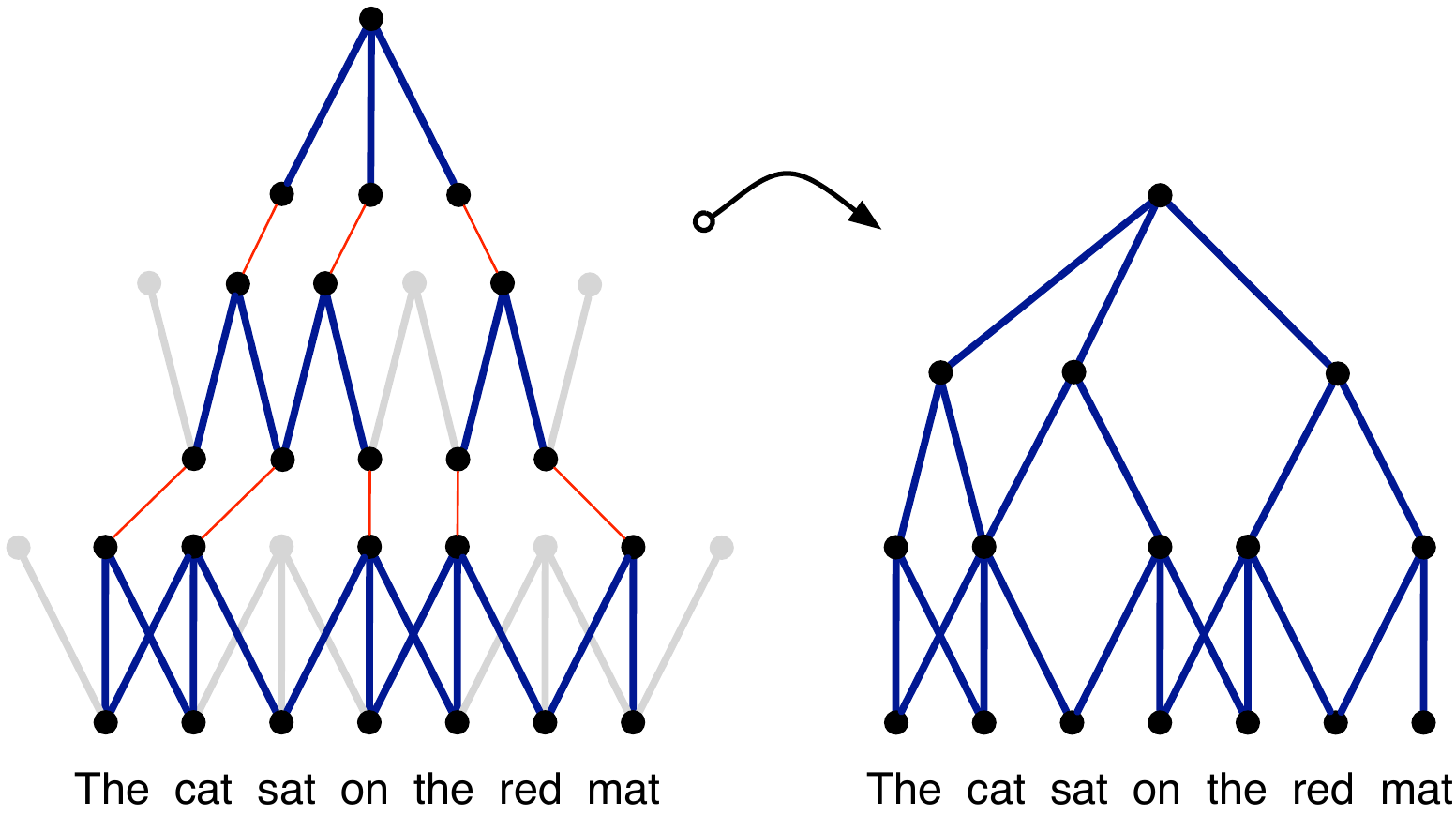}}
\vspace{-0.7cm}
\caption{Subgraph of a feature graph induced over an input sentence in a Dynamic Convolutional Neural Network. The full induced graph has multiple subgraphs of this kind with a distinct set of edges; subgraphs may merge at different layers. The left diagram emphasises the pooled nodes. The width of the convolutional filters is 3 and 2 respectively. With dynamic pooling, a filter with small width at the higher layers can relate phrases far apart in the input sentence.}
\vspace{-0.3cm}
\end{figure}

Various types of models of meaning have been proposed. Composition based methods have been applied to vector representations of word meaning obtained from co-occurrence statistics to obtain vectors for longer phrases. In some cases, composition is defined by algebraic operations over word meaning vectors to produce sentence meaning vectors \cite{Erk2008,mitchell2008vector,DBLP:journals/cogsci/MitchellL10,turney2012domain,erk2012vector,clarke2012context}. In other cases, a composition function is learned and either tied to particular syntactic relations \cite{Guevara2010,zanzotto2010estimating} or to particular word types \cite{DBLP:conf/emnlp/BaroniZ10,Coecke2010,grefenstette2011experimental,karts_sadr_emnlp,grefenstette2013category}. Another approach represents the meaning of sentences by way of automatically extracted logical forms \cite{DBLP:conf/uai/ZettlemoyerC05}. 

A central class of models are those based on neural networks. These range from basic neural bag-of-words or bag-of-$n$-grams models to the more structured recursive neural networks and to time-delay neural networks based on convolutional operations \cite{collobert:2008,SocherEtAl2011:RAE,KalchbrennerBlunsom2013:RCNN}. Neural sentence models have a number of advantages. They can be trained to obtain generic vectors for words and phrases by predicting, for instance, the contexts in which the words and phrases occur. Through supervised training, neural sentence models can fine-tune these vectors to information that is specific to a certain task. Besides comprising powerful classifiers as part of their architecture, neural sentence models can  be used to condition a neural language model to generate sentences word by word \cite{schwenk2012cstm,DBLP:conf/slt/MikolovZ12,kalchbrenner13emnlp}.

We define a convolutional neural network architecture and apply it to the semantic modelling of sentences. The network handles input sequences of varying length. The layers in the network interleave one-dimensional convolutional layers and dynamic $k$-max pooling layers. Dynamic $k$-max pooling is a generalisation of the max pooling operator. The max pooling operator is a non-linear  subsampling function that returns the maximum of a set of values \cite{lecun-98}. The operator is generalised in two respects. First, $k$-max pooling over a linear sequence of values returns the subsequence of $k$ maximum values in the sequence, instead of the single maximum value. Secondly, the pooling parameter $k$ can be dynamically chosen by making $k$ a function of other aspects of the network or the input. 

The convolutional layers apply one-dimensional filters across each row of features in the sentence matrix. Convolving the same filter with the $n$-gram at every position in the sentence allows the features to be extracted independently of their position in the sentence. A convolutional layer followed by a dynamic pooling layer and a non-linearity form a feature map. Like in the convolutional networks for object recognition \cite{lecun-98}, we enrich the representation in the first layer by computing multiple feature maps with different filters applied to the input sentence. Subsequent layers also have multiple feature maps computed by convolving filters with all the maps from the layer below. The weights at these layers form an order-4 tensor. The resulting architecture is dubbed a Dynamic Convolutional Neural Network.

Multiple layers of convolutional and dynamic pooling operations induce a structured feature graph over the input sentence. Figure 1 illustrates such a graph. Small filters at higher layers can capture syntactic or semantic relations between non-continuous phrases that are far apart in the input sentence. The feature graph induces  a hierarchical structure somewhat akin to that in a syntactic parse tree. The structure is not tied to purely syntactic relations and is internal to the neural network.

We experiment with the network in four settings. The first two experiments involve predicting the sentiment of movie reviews \cite{Socher-etal:2013}. The network outperforms other approaches in both the binary and the multi-class experiments. The third experiment involves the categorisation of questions in six question types in the TREC dataset \cite{li2002learning}. The network matches the accuracy of other state-of-the-art methods that are based on large sets of engineered features and hand-coded knowledge resources. 
The fourth experiment involves predicting the sentiment of Twitter posts using distant supervision \cite{Go_Bhayani_Huang_2009}. The network is trained on 1.6 million tweets labelled automatically according to the emoticon that occurs in them. On the hand-labelled test set, the network achieves a greater than $25\%$ reduction in the prediction error with respect to the strongest unigram and bigram baseline reported in \newcite{Go_Bhayani_Huang_2009}.

The outline of the paper is as follows. Section 2 describes the background to the DCNN including central concepts and related neural sentence models. Section 3 defines the relevant operators and the layers of the network. Section 4 treats of the induced feature graph and other properties of the network. Section 5 discusses the experiments and inspects the learnt feature detectors.\footnote{Code available at \url{www.nal.co}}

\section{Background}
\label{back}

The layers of the DCNN are formed by a convolution operation followed by a pooling operation. We begin with a review of related neural sentence models. Then we describe the operation of \emph{one-dimensional convolution} and the classical Time-Delay Neural Network (TDNN) \cite{DBLP:journals/ai/Hinton89,Waibel:1990:PRU:108235.108263}. By adding a max pooling layer to the network, the TDNN can be adopted as a sentence model \cite{collobert:2008}.

\subsection{Related Neural Sentence Models}
Various neural sentence models have been described. A general class of basic sentence models is that of Neural Bag-of-Words (NBoW) models. These generally consist of a projection layer that maps words, sub-word units or $n$-grams  to high dimensional embeddings; the latter are then combined component-wise with an operation such as summation. The resulting combined vector is  classified through one or more fully connected layers.

A model that adopts a more general structure provided by an external parse tree is the Recursive Neural Network (RecNN) \cite{pollack:recursive,DBLP:conf/ki/KuchlerG96,SocherEtAl2011:RAE,hermann-blunsom:2013:ACL2013}. At every node in the tree the contexts at the left and right children of the node are combined by a classical layer. The weights of the layer are shared across all nodes in the tree. The layer computed at the top node gives a representation for the sentence. 
The Recurrent Neural Network (RNN) is a special case of the recursive network where the structure that is followed is a simple linear chain \cite{journals/tnn/GersS01,DBLP:conf/icassp/MikolovKBCK11}. The RNN is primarily used as a language model, but may also be viewed as a sentence model with a linear structure. The layer computed at the last word represents the sentence. 

Finally, a further class of neural sentence models is based on the convolution operation and the TDNN architecture \cite{collobert:2008,KalchbrennerBlunsom2013:RCNN}. Certain concepts used in these models are central to the DCNN and we describe them next.

\subsection{Convolution}
\label{conv_types_sect}
The \emph{one-dimensional convolution} is an operation between a vector of weights $\vec{m}\in\mathbb{R}^m$ and a vector of inputs viewed as a sequence $\vec{s} \in \mathbb{R}^s$. The vector $\vec{m}$ is the \emph{filter} of the convolution. Concretely, we think of $\vec{s}$ as the input sentence and $\vec{s}_i \in \mathbb{R}$ is a single feature value associated with the $i$-th word in the sentence. The idea behind the one-dimensional convolution is to take the dot product of the vector $\vec{m}$ with each $m$-gram in the sentence $\vec{s}$ to obtain another sequence $\vec{c}$:

\vspace{-7pt}
\begin{equation} 
\label{conv_f}
\vec{c}_j  = \vec{m}^{\intercal} \vec{s}_{j-m+1:j}
\end{equation} 

\vspace{-2pt}
\noindent
Equation \ref{conv_f} gives rise to two types of convolution depending on the range of the index $j$. The \emph{narrow} type of convolution requires that $s \geq m$ and yields a sequence $\vec{c} \in \mathbb{R}^{s-m+1}$ with $j$ ranging from $m$ to $s$. The \emph{wide} type of convolution does not have requirements on $s$ or $m$ and yields a sequence $\vec{c} \in \mathbb{R}^{s+m-1}$ where the index $j$ ranges from $1$ to $s+m-1$.  Out-of-range input values $\vec{s}_i$ where $i<1$ or $i>s$ are taken to be zero. The result of the narrow convolution is a subsequence of the result of the wide convolution. The two types of one-dimensional convolution are illustrated in Fig.~2.

The trained weights in the filter $\vec{m}$ correspond to a linguistic feature detector that learns to recognise a specific class of $n$-grams. These $n$-grams have size $n\leq m$, where $m$ is the width of the filter. Applying the weights $\vec{m}$ in a wide convolution has some advantages over applying them in a narrow one. A wide convolution ensures that all weights in the filter reach the entire sentence, including the words at the margins. This is particularly significant when $m$ is set to a relatively large value such as 8 or 10. In addition, a wide convolution guarantees that the application of the filter $\vec{m}$ to the input sentence $\vec{s}$ always produces a valid non-empty result $\vec{c}$, independently of the width $m$ and the sentence length $s$. We next describe the classical convolutional layer of a TDNN.

\subsection{Time-Delay Neural Networks}
\label{tdnn}

\begin{figure}
\centering
\label{conv_types}
{\includegraphics[width=0.48\textwidth]{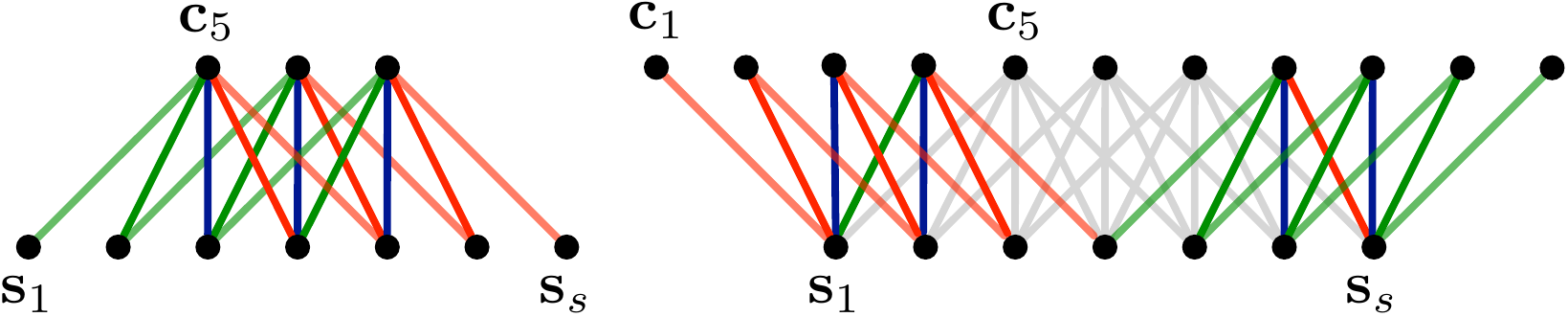}}
\vspace{-0.7cm}
\caption{Narrow and wide types of convolution. The filter $\vec{m}$ has size $m=5$. }
\vspace{-0.5cm}
\end{figure}

A TDNN convolves a sequence of inputs $\vec{s}$ with a set of weights $\vec{m}$. As in the TDNN for phoneme recognition  \cite{Waibel:1990:PRU:108235.108263}, the sequence $\vec{s}$ is viewed as having a time dimension and the convolution is applied over the time dimension. Each $\vec{s}_j$ is often not just a single value, but a vector of $d$ values so that $\vec{s}\in\mathbb{R}^{d\times s}$. Likewise, $\vec{m}$ is a matrix of weights of size ${d \times m}$. Each row of $\vec{m}$ is convolved with the corresponding row of $\vec{s}$ and the convolution is usually of the narrow type. Multiple convolutional layers may be stacked by taking the resulting sequence $\vec{c}$ as  input to the next layer. 

The Max-TDNN sentence model  is based on the architecture of a TDNN \cite{collobert:2008}. In the model, a convolutional layer of the narrow type is applied to the sentence matrix $\vec{s}$, where each column corresponds to the feature vector $\vec{w}_i\in\mathbb{R}^d$ of a word in the sentence:
\begin{equation}
\label{sentmat}
\vec{s} = \begin{bmatrix}
\kern.6em\vline & \kern.2em\vline\kern.2em & \vline\kern.6em \\
\vec{w}_1 & \hdots & \vec{w}_s \\
\kern.6em\vline & \kern.2em\vline\kern.2em & \vline\kern.6em \\
\end{bmatrix}
\end{equation}
 To address the problem of varying sentence lengths, the Max-TDNN takes the maximum of each row in the resulting matrix $\vec{c}$ yielding a vector of $d$ values:
\begin{equation}
\vec{c}_{max} = \begin{bmatrix}
\max(\vec{c}_{1,:}) \\
\vdots \\
\max(\vec{c}_{d,:}) \\
\end{bmatrix}
\end{equation}
The aim is to capture the most relevant feature, i.e. the one with the highest value, for each of the $d$ rows of the resulting matrix $\vec{c}$. The fixed-sized vector $\vec{c}_{max}$ is then used as input to a fully connected layer for  classification.

The Max-TDNN model has many desirable properties. It is sensitive to the order of the words in the sentence and it does not depend on external language-specific features such as dependency or constituency parse trees. It also gives largely uniform importance to the signal coming from each of the words in the sentence, with the exception of words at the margins that are considered fewer times in the computation of the narrow convolution. But the model also has some limiting aspects. The range of the feature detectors is limited to the span $m$ of the weights. Increasing $m$ or stacking multiple convolutional layers of the narrow type makes the range of the feature detectors larger; at the same time it also exacerbates the neglect of the margins of the sentence and increases the minimum size $s$ of the input sentence required by the convolution. For this reason higher-order and long-range feature detectors cannot be easily incorporated into the model. 
\begin{figure}
\label{cnnsm}
\vspace{-0.8cm}
{\includegraphics[width=0.5\textwidth]{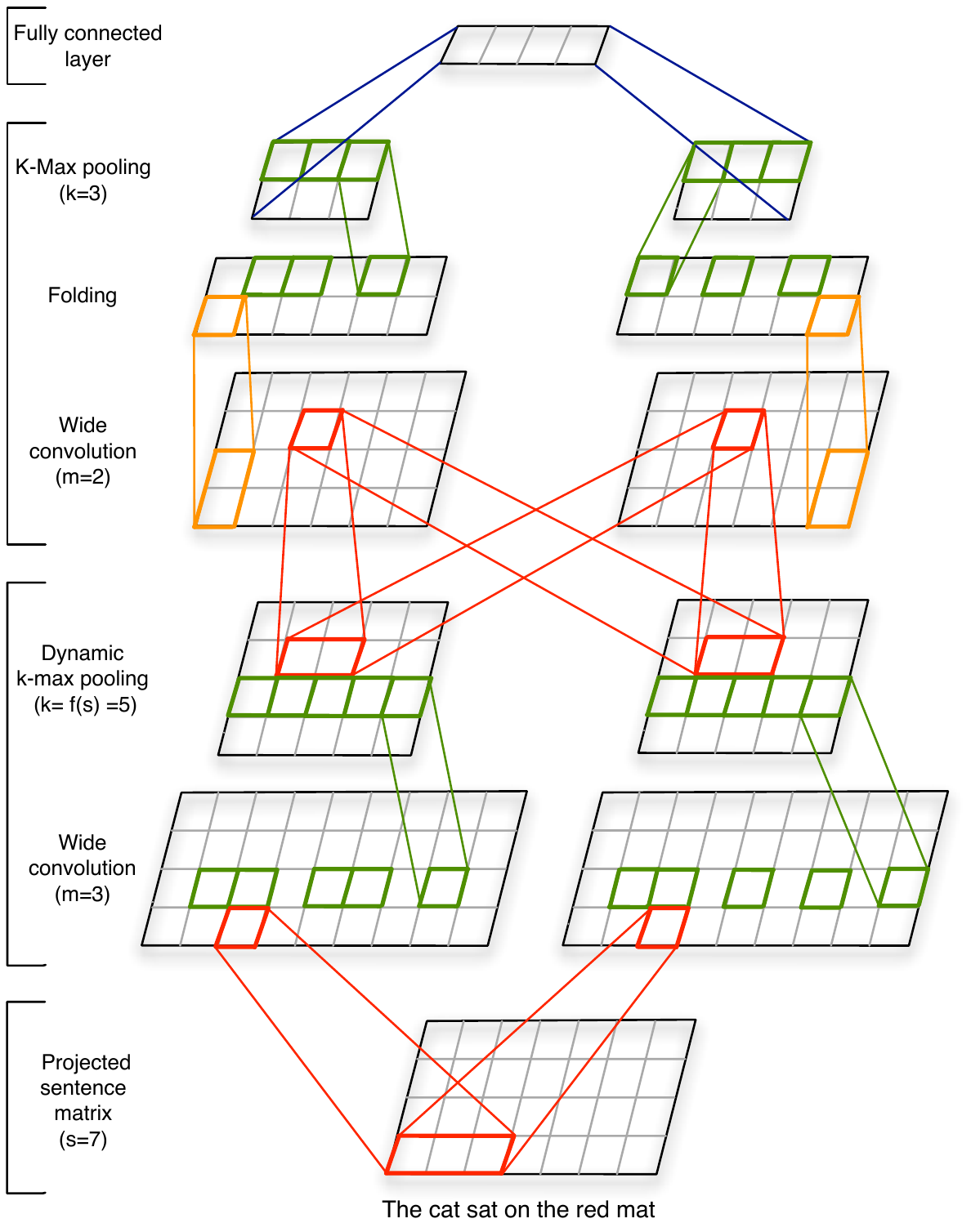}}
\caption{A DCNN for the seven word input sentence. Word embeddings have size $d=4$. The network has two convolutional layers with two feature maps each. The widths of the filters at the two layers are respectively 3 and 2. The (dynamic) $k$-max pooling layers have values $k$ of $5$ and 3.   }
\vspace{-0.3cm}
\end{figure}
The max pooling operation has some disadvantages too. It cannot distinguish whether a relevant feature in one of the rows occurs just one or multiple times and it forgets the order in which the features occur. More generally, the pooling factor by which the signal of the matrix is reduced at once corresponds to $s-m+1$; even for moderate values of $s$ the pooling factor can be excessive.
The aim of the next section is to address these limitations while preserving the  advantages.

\section{Convolutional Neural Networks with Dynamic $k$-Max Pooling}
\label{convdyn}

We model sentences using a convolutional architecture that alternates wide convolutional layers with dynamic pooling layers given by  \emph{dynamic $k$-max pooling}. In the network the width of a feature map at an intermediate layer varies depending on the length of the input sentence; the resulting architecture is the Dynamic Convolutional Neural Network. Figure 3 represents a DCNN. We proceed to describe the network in detail.

\subsection{Wide Convolution}

Given an input sentence, to obtain the first layer of the DCNN we take the embedding $\vec{w}_i \in \mathbb{R}^d$ for each word in the sentence and construct the sentence matrix $\vec{s} \in \mathbb{R}^{d\times s}$ as in Eq.~\ref{sentmat}. The values in the embeddings $\vec{w}_i$ are parameters that are optimised during training. 
A convolutional layer in the network is obtained by convolving a matrix of weights $\vec{m} \in \mathbb{R}^{d \times m}$ with the matrix of activations at the layer below. For example, the second layer is obtained by applying a convolution to the sentence matrix $\vec{s}$ itself.  Dimension $d$ and filter width $m$ are hyper-parameters of the network. We let the operations be \emph{wide} one-dimensional convolutions as described in Sect.~\ref{conv_types_sect}. The resulting matrix $\vec{c}$ has dimensions $d \times (s+m-1)$. 

\subsection{$k$-Max Pooling}
\label{kmax}
We next describe a pooling operation that is a generalisation of the max pooling over the time dimension used in the Max-TDNN sentence model and different from the local max pooling operations applied in a convolutional network for object recognition \cite{lecun-98}. Given a value $k$ and a sequence $\vec{p} \in \mathbb{R}^p$  of length $p \geq k$, \emph{$k$-max pooling} selects the subsequence $\vec{p}^k_{max}$ of the $k$ highest values of $\vec{p}$. The order of the values in  $\vec{p}^k_{max}$ corresponds to their original order in $\vec{p}$. 

The $k$-max pooling operation makes it possible to pool the $k$ most active features in $\vec{p}$ that may be a number of positions apart; it preserves the order of the features, but is insensitive to their specific positions. It can also discern more finely the number of times the feature is highly activated in $\vec{p}$ and the progression by which the high activations of the feature change across $\vec{p}$. 
The $k$-max pooling operator is applied in the network after the topmost convolutional layer. This guarantees that the input to the fully connected layers is independent of the length of the input sentence.
But, as we see next, at intermediate convolutional layers the pooling parameter $k$ is not fixed, but is dynamically selected in order to allow for a smooth extraction of higher-order and longer-range features.

\subsection{Dynamic $k$-Max Pooling}

A \emph{dynamic $k$-max pooling} operation is a $k$-max pooling operation where we let $k$ be a function of the length of the sentence and the depth of the network. Although many functions are possible, we simply model the pooling parameter as follows:

\vspace{-3pt}
\begin{equation}
\label{pool}
k_l = \max(\ k_{top},\ \lceil \frac{L - l}{L} s \rceil \ )
\end{equation}

\vspace{-3pt}
\noindent
where $l$ is the number of the current convolutional layer to which the pooling is applied and $L$ is the total number of convolutional layers in the network; $k_{top}$ is the fixed pooling parameter for the topmost convolutional layer (Sect.~\ref{kmax}). For instance, in a network with three convolutional layers and $k_{top} = 3$, for an input sentence of length $s=18$, the pooling parameter at the first layer is $k_1 = 12$ and the pooling parameter at the second layer is $k_2 = 6$; the third layer has the fixed pooling parameter $k_3 = k_{top} = 3$. Equation \ref{pool} is a model of the number of values needed to describe the relevant parts of the progression of an $l$-th order feature over a sentence of length $s$. For an example in sentiment prediction, according to the equation a first order feature such as a positive word occurs \emph{at most} $k_1$ times in a sentence of length $s$, whereas a second order feature such as a negated phrase or clause occurs at most $k_2$ times.

\subsection{Non-linear Feature Function}

After (dynamic) $k$-max pooling is applied to the result of a convolution, a bias $\vec{b} \in \mathbb{R}^d$ and a non-linear function $g$ are applied component-wise to the pooled matrix. There is a single bias value for each row of the pooled matrix. 

If we temporarily ignore the pooling layer, we may state how one computes each $d$-dimensional column $a$ in the matrix $\vec{a}$ resulting after the convolutional and non-linear layers. Define $\vec{M}$ to be the matrix of diagonals:
\vspace{-3pt}
\begin{equation}
\label{mmm}
\vec{M} =  \left[\mbox{diag}(\vec{m}_{:,1}  ), \hdots,\mbox{diag}(\vec{m}_{:,m} )\right]
\end{equation}

\vspace{-3pt}
\noindent where $\vec{m}$ are the weights of the $d$ filters of the wide convolution.
Then after the first pair of a convolutional and a non-linear layer, each column $a$ in the matrix $\vec{a}$ is obtained as follows, for some index $j$:
\vspace{-3pt}
\begin{equation}
\label{form}
a = g \left(\vec{M}\left[ \begin{array}{ccc}
\vec{w}_{j} \\
\vdots \\
\vec{w}_{j+m-1}  \end{array} \right] + \vec{b} \right)
\end{equation}

\vspace{-3pt}
\noindent Here $a$ is a column of first order features. Second order features are similarly obtained by applying Eq.~\ref{form} to a sequence of first order features $a_j,...,a_{j+m'-1}$ with another weight matrix $\vec{M}'$. Barring pooling, Eq.~\ref{form} represents a core aspect of the feature extraction function and has a rather general form that we return to below. Together with pooling, the feature function induces position invariance and makes the range of higher-order features variable.

\subsection{Multiple Feature Maps}

So far we have described  how one applies a wide convolution, a (dynamic) $k$-max pooling layer and a non-linear function to the input sentence matrix to obtain a first order \emph{feature map}. The three operations can be repeated to yield feature maps of increasing order and a network of increasing depth. We denote a feature map of the $i$-th order by $\vec{F}^i$. As in convolutional networks for object recognition, to increase the number of learnt feature detectors of a certain order, multiple feature maps $\vec{F}^i_1, \hdots, \vec{F}^i_n$ may be computed in parallel at the same layer. Each feature map $\vec{F}^i_j$ is computed by convolving a distinct set of filters arranged in a matrix $\vec{m}^i_{j,k}$ with each feature map $\vec{F}^{i-1}_k$ of the lower order $i-1$ and summing the results:
\vspace{-8pt}
\begin{equation}
\vec{F}^{i}_j = \sum_{k=1}^n \vec{m}^i_{j,k} *\vec{F}^{i-1}_{k}
\end{equation}
\vspace{-12pt}

\noindent where $*$ indicates the wide convolution. The weights $\vec{m}^i_{j,k}$ form an order-4 tensor. After the wide convolution, first dynamic $k$-max pooling and then the non-linear function are applied individually to each map.

\subsection{Folding}
In the formulation of the network so far, feature detectors applied to an individual row of the sentence matrix $\vec{s}$ can have many orders and create complex dependencies across the same rows in multiple feature maps. Feature detectors in different rows, however, are independent of each other until the top  fully connected layer. Full dependence between different rows could be achieved by making $\vec{M}$ in Eq.~\ref{mmm} a full matrix instead of a sparse matrix of diagonals. Here we explore a simpler method called \emph{folding} that does not introduce any additional parameters. After a convolutional layer and before (dynamic) $k$-max pooling, one just sums every two rows in a feature map component-wise. For a map of $d$ rows, folding returns a map of $d/2$ rows, thus halving the size of the representation. With a folding layer, a feature detector of the $i$-th order depends now on two rows of feature values in the lower maps of order $i-1$. This ends the description of the DCNN.

\section{Properties of the Sentence Model}
\label{prop}

We describe some of the properties of the sentence model based on the DCNN. We describe the  notion of the \emph{feature graph} induced over a sentence by the succession of convolutional and pooling layers. We briefly relate the properties to those of other neural sentence models.

\subsection{Word and $n$-Gram Order}
One of the basic properties is sensitivity to the order of the words in the input sentence. For most applications and in order to learn fine-grained feature detectors, it is beneficial for a model to be able to discriminate whether a specific $n$-gram occurs in the input. Likewise, it is beneficial for a model to be able to tell the \emph{relative} position of the most relevant $n$-grams. The network is designed to capture these two aspects. The filters $\vec{m}$ of the wide convolution in the first layer can learn to recognise specific $n$-grams that have size less or equal to the filter width $m$; as we see in the experiments, $m$ in the first layer is often set to a relatively large value such as $10$. The subsequence of $n$-grams extracted by the generalised pooling operation induces invariance to absolute positions, but maintains their order and relative positions.

As regards the other neural sentence models, the class of NBoW models is by definition insensitive to word order. A sentence model based on a recurrent neural network is sensitive to word order, but it has a bias towards the latest words that it takes as input \cite{DBLP:conf/icassp/MikolovKBCK11}. This gives  the RNN excellent performance at language modelling, but it is suboptimal for remembering at once the $n$-grams further back in the input sentence. Similarly, a recursive neural network is sensitive to word order but has a  bias towards the topmost nodes in the tree; shallower trees mitigate this effect to some extent \cite{SocherEtAl2013:DTRNN}. As seen in Sect.~\ref{tdnn}, the Max-TDNN is sensitive to word order, but max pooling only picks out a single $n$-gram feature in each row of the sentence matrix. 

\subsection{Induced Feature Graph}

Some sentence models use internal or external structure to compute the representation for the input sentence. In a DCNN, the convolution and pooling layers induce an internal feature  graph over the input. A node from a layer is connected to a node from the next higher layer if the lower node is involved in the convolution that computes the value of the higher node. Nodes that are not selected by the pooling operation at a layer are dropped from the graph. After the last pooling layer, the remaining nodes connect to a single topmost root.  The induced graph is a connected, directed acyclic graph with weighted edges and a root node; two equivalent representations of an induced graph are given in Fig.~\ref{DAGs}. In a DCNN without folding layers, each of the $d$ rows of the sentence matrix induces a subgraph that joins the other subgraphs only at the root node. Each  subgraph may have a different shape that reflects the kind of relations that are detected in that subgraph. The effect of folding layers is to join pairs of subgraphs at lower layers before the top root node. 

Convolutional networks for object recognition also induce a feature graph over the input image. What makes the feature graph of a DCNN peculiar is the global range of the pooling operations. The (dynamic) $k$-max pooling operator can draw together features that correspond to words that are many positions apart in the sentence. Higher-order features have highly variable ranges that can be either short and focused or global and long as the input sentence. Likewise, the edges of a subgraph in the induced graph reflect these varying ranges. The subgraphs can either be localised to one or more parts of the sentence or spread more widely across the sentence. This structure is internal to the network and is defined by the forward propagation of the input through the network. 

Of the other sentence models, the NBoW is a shallow model and the RNN has a linear chain structure. The subgraphs induced in the Max-TDNN model have a single fixed-range feature obtained through max pooling. The recursive neural network follows the structure of an external  parse tree. Features of variable range are computed at each node of the tree combining one or more of the children of the tree. Unlike in a DCNN, where one learns a clear hierarchy of feature orders, in a RecNN low order features like those of single words can be directly combined with higher order features computed from entire clauses. A DCNN generalises many of the structural aspects of a RecNN. The feature extraction function as stated in Eq.~\ref{form} has a more general form than that in a RecNN, where the value of $m$ is generally 2. Likewise, the induced graph structure in a DCNN is more general than a parse tree in that it is not limited to syntactically dictated phrases; the graph structure can capture short or long-range semantic relations between words that do not necessarily correspond to the syntactic relations in a parse tree. The DCNN has internal input-dependent structure and does not rely on externally provided parse trees, which makes the DCNN directly applicable to hard-to-parse sentences such as tweets and to sentences from any language.

\section{Experiments}
\label{exp}
\begin{table}
\label{exp:sent}
\centering
\small
\begin{tabular}{ l  c  c  c  c  c c  } 
\toprule
 Classifier &  Fine-grained (\%) & Binary (\%) \\
\midrule
\textsc{NB}  & 41.0 & 81.8 \\ \midrule
\textsc{BiNB} & 41.9 & 83.1 \\ \midrule
\textsc{SVM}   & 40.7 & 79.4 \\ \midrule
\textsc{RecNTN}&  45.7 & 85.4 \\ \midrule 
\textsc{Max-TDNN} & 37.4 & 77.1 \\ \midrule
\textsc{NBoW} & 42.4 &  80.5 \\ \midrule
\textsc{DCNN} & \textbf{48.5} & \textbf{86.8} \\ \midrule
\bottomrule
\end{tabular}
\caption{Accuracy of sentiment prediction in the movie reviews dataset. The first four results are reported from \newcite{Socher-etal:2013}. The baselines \textsc{NB} and \textsc{BiNB} are Naive Bayes classifiers with, respectively, unigram features and unigram and bigram features. \textsc{SVM} is a support vector machine with unigram and bigram features. \textsc{RecNTN} is a recursive neural network with a tensor-based feature function, which relies on external structural features given by a parse tree and performs best among the RecNNs. }
\vspace{-0.5cm}
\end{table}

We test the network on four different experiments. We begin by specifying aspects of the implementation and the training of the network. We then relate the results of the experiments and we inspect the learnt feature detectors.

\subsection{Training}

In each of the experiments, the top layer of the network has a fully connected layer followed by a softmax non-linearity that predicts the probability distribution over classes given the input sentence.
The network is trained to minimise the cross-entropy of the predicted and true distributions; the objective includes an $L_2$ regularisation term over the parameters.
The set of parameters comprises the word embeddings, the filter weights and the weights from the fully connected layers. The network is trained with mini-batches by backpropagation and the gradient-based optimisation is performed using the Adagrad update rule \cite{duchi2011adagrad}. 
Using the well-known convolution theorem, we can compute fast one-dimensional linear convolutions at all rows of an input matrix by using Fast Fourier Transforms. To exploit the parallelism of the operations, we train the network on a GPU. A Matlab implementation processes multiple millions of input sentences per hour on one GPU, depending primarily on the number of layers used in the network. 

\subsection{Sentiment Prediction in Movie Reviews}
\label{sec:sent}

The first two experiments concern the prediction of the sentiment of movie reviews in the Stanford Sentiment Treebank \cite{Socher-etal:2013}. The output variable is binary in one experiment and can have five possible outcomes in the other: negative, somewhat negative, neutral, somewhat positive, positive. In the binary case, we use the given splits of 6920 training, 872 development and 1821 test sentences. Likewise, in the fine-grained case, we use the standard 8544/1101/2210 splits.   Labelled phrases that occur as subparts of the training sentences are treated as independent training instances. The size of the vocabulary is 15448. 

\begin{table}
\label{sec:qc}
\small
\centering
\begin{tabular}{ l  l c  c  c  c  c c  } 
\toprule
 Classifier & Features & Acc. (\%) \\ \midrule
\multirow{2}{*}{\textsc{Hier}} & unigram, POS, head chunks & 91.0  \\ 
 & NE, semantic relations \\\midrule
\multirow{3}{*}{\textsc{MaxEnt}}  &  unigram, bigram, trigram & 92.6  \\ 
&POS, chunks, NE, supertags & \\
&CCG parser,  WordNet & \\\midrule
\multirow{4}{*}{\textsc{MaxEnt}}  & unigram, bigram, trigram & 93.6  \\ 
& POS, wh-word, head word \\
&  word shape, parser  \\
& hypernyms, WordNet\\\midrule 
\multirow{4}{*}{\textsc{SVM}} & unigram, POS, wh-word & 95.0 \\ 
& head word, parser \\
& hypernyms, WordNet\\
& 60 hand-coded rules \\\midrule 
\textsc{Max-TDNN} & unsupervised vectors & 84.4 \\ \midrule
\textsc{NBoW} & unsupervised  vectors & 88.2 \\ \midrule
\textsc{DCNN} & unsupervised vectors & {93.0} \\ \midrule
\bottomrule
\end{tabular}
\vspace{-0.1cm}
\caption{Accuracy of six-way question classification on the TREC questions dataset. The second column details the external features used in the various approaches. The first four results are respectively from \newcite{li2002learning}, \newcite{blunsom06question}, \newcite{Huang:2008:QCU:1613715.1613835} and \newcite{silva2011}. }
\vspace{-0.5cm}
\end{table}

Table 1 details the results of the experiments. In the three neural sentence models---the Max-TDNN, the NBoW and the DCNN---the word vectors are parameters of the models that are randomly initialised; their dimension $d$ is set to 48. The Max-TDNN has a filter of width $6$ in its narrow convolution at the first layer; shorter phrases are padded with zero vectors. The convolutional layer is followed by a non-linearity, a max-pooling layer and a softmax classification layer. The NBoW sums the word vectors and applies a non-linearity followed by a softmax classification layer.  The adopted non-linearity is the $\tanh$ function.
The hyper parameters of the DCNN are as follows. The binary result is based on a DCNN that has a wide convolutional layer followed by a folding layer, a dynamic $k$-max pooling layer and a non-linearity; it has a second wide convolutional layer followed by a folding layer, a $k$-max pooling layer and a non-linearity.  The width of the convolutional filters is 7 and 5, respectively. The value of $k$ for the top $k$-max pooling is 4. The number of feature maps at the first convolutional layer is 6; the number of maps at the second convolutional layer is 14. The network is topped by a softmax classification layer. The DCNN for the fine-grained result has  the same architecture, but the filters have size 10 and 7, the top pooling parameter $k$ is 5 and the number of maps is, respectively, 6 and 12. The networks use the $\tanh$ non-linear function. At training time we apply dropout to the penultimate layer after the last $\tanh$ non-linearity \cite{DBLP:journals/corr/abs-1207-0580}. 
\begin{table}
\label{sec:twitter}
\centering
\small
\begin{tabular}{ l  c  c  c  c  c c  } 
\toprule
 Classifier & Accuracy (\%) \\
\midrule
\textsc{SVM}  & 81.6 \\ \midrule
\textsc{BiNB}  & 82.7  \\ \midrule
\textsc{MaxEnt} & 83.0 \\ \midrule
\textsc{Max-TDNN} & {78.8} \\ \midrule
\textsc{NBoW} & {80.9} \\ \midrule
\textsc{DCNN} & \textbf{87.4} \\ \midrule
\bottomrule
\end{tabular}
\vspace{-0.1cm}
\caption{Accuracy on the Twitter sentiment dataset. The three non-neural classifiers are based on unigram and bigram features; the results are reported from \cite{Go_Bhayani_Huang_2009}. }
\vspace{-0.5cm}
\end{table}

\begin{figure*}
\centering
\label{neurons}
{\includegraphics[width=\textwidth]{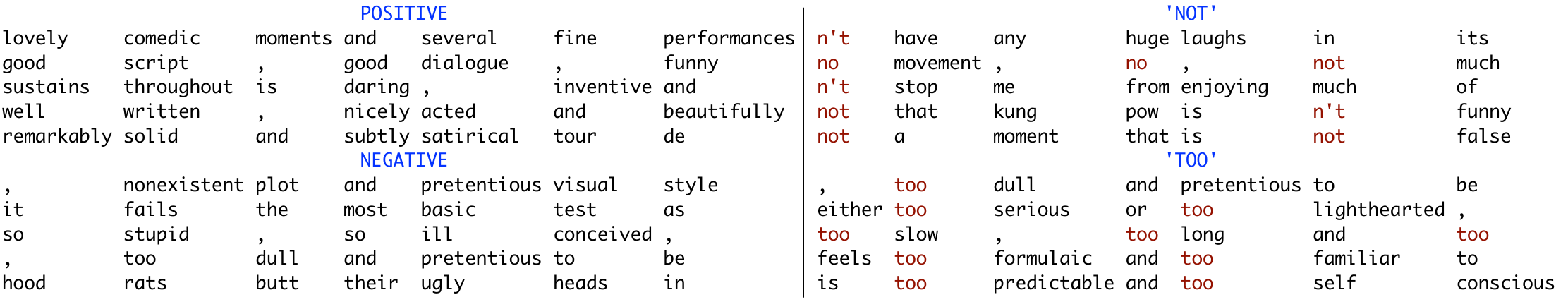}}
\vspace{-0.8cm}
\caption{Top five $7$-grams at four feature detectors in the first layer of the network.}
\vspace{-0.5cm}
\end{figure*}
We see that the DCNN significantly outperforms the other neural and non-neural models. The NBoW performs similarly to the non-neural $n$-gram based classifiers. The Max-TDNN performs worse than the NBoW likely due to the excessive pooling of the max pooling operation; the latter discards most of the sentiment features of the words in the input sentence.
Besides the RecNN that uses an external parser to produce structural features for the model, the other models use $n$-gram based or neural features that do not require external resources or additional annotations. In the next experiment we compare the performance of the DCNN with those of methods that use heavily engineered resources.

\subsection{Question Type Classification}

As an aid to question answering, a question may be classified as belonging to one of many question types. The TREC questions dataset involves six different question types, e.g. whether the question is about a location, about a person or about some numeric information \cite{li2002learning}.  The training dataset consists of 5452 labelled questions whereas the test dataset consists of 500 questions. 

The results are reported in Tab.~2. The non-neural approaches use a classifier over a large number of manually engineered features and hand-coded resources.  For instance, \newcite{blunsom06question} present a Maximum Entropy model that relies on 26 sets of syntactic and semantic features including unigrams, bigrams, trigrams, POS tags, named entity tags, structural relations from a CCG parse and WordNet synsets. 
We evaluate the three neural models on this dataset with mostly the same hyper-parameters as in the binary sentiment experiment of Sect.~\ref{sec:sent}. As the dataset is rather small, we use lower-dimensional word vectors with $d=32$ that are initialised with embeddings trained in an unsupervised way to predict contexts of occurrence \cite{turian2010word}. The DCNN uses a single convolutional layer with filters of size 8 and 5 feature maps. The difference between the performance of the DCNN and that of the other high-performing methods in Tab. 2 is not significant ($p <0.09$). Given that the only labelled information used to train the network is the training set itself, it is notable that the network matches the performance of state-of-the-art classifiers that rely on large amounts of engineered features and rules and hand-coded resources.

\subsection{Twitter Sentiment Prediction with Distant Supervision}

In our final experiment, we train the models on a large dataset of tweets, where a tweet is automatically labelled as positive or negative depending on the emoticon that occurs in it. The training set consists of 1.6 million tweets with emoticon-based labels and the test set of about 400 hand-annotated tweets. We preprocess the tweets minimally following the procedure described in \newcite{Go_Bhayani_Huang_2009}; in addition, we also lowercase all the  tokens. This results in a vocabulary of 76643 word types. The architecture of the DCNN and of the other neural models is the same as the one used in the binary experiment of Sect.~\ref{sec:sent}. The randomly initialised word embeddings are increased in length to a dimension of $d=60$. Table 3 reports the results of the experiments. We see a significant increase in the performance of the DCNN with respect to the non-neural $n$-gram based classifiers; in the presence of large amounts of training data these classifiers constitute particularly strong baselines. We see that the ability to train a sentiment classifier on automatically extracted emoticon-based labels extends to the DCNN and results in highly accurate performance. The difference in performance between the DCNN and the NBoW further suggests that the ability of the DCNN to both capture features based on long $n$-grams and to hierarchically combine these features is highly beneficial.

\subsection{Visualising Feature Detectors}

A filter in the DCNN is associated with a feature detector or neuron that learns during training to be particularly active when presented with a specific sequence of input words. In the first layer, the sequence is a continuous $n$-gram from the input sentence; in higher layers, sequences can be made of multiple separate $n$-grams. We visualise the feature detectors in the first layer of the network trained on the binary sentiment task (Sect.~\ref{sec:sent}). Since the filters have width 7, for each of the 288 feature detectors we rank all $7$-grams occurring in the validation and test sets according to their activation of the detector. Figure \ref{neurons} presents the top five $7$-grams for four feature detectors. Besides the expected detectors for positive and negative sentiment, we find detectors for particles such as `not' that negate sentiment and such as `too' that potentiate sentiment. We find detectors for multiple other notable constructs including `all', `or', `with...that', `as...as'. The feature detectors learn to recognise not just single $n$-grams, but patterns within $n$-grams that  have syntactic, semantic or structural significance.

\vspace{-3pt}
\section{Conclusion}

We have described a dynamic convolutional neural network that uses the dynamic $k$-max pooling operator as a non-linear subsampling function. The feature graph induced by the network  is able to capture word relations of varying size. The network achieves high performance on question and sentiment classification without requiring external features  as provided by parsers or other resources.

\section*{Acknowledgements}
We thank Nando de Freitas and Yee Whye Teh for great discussions on the paper. This work was supported by a Xerox Foundation Award, EPSRC grant number EP/F042728/1, and EPSRC grant number EP/K036580/1.

\bibliographystyle{acl}
\bibliography{Master}

\end{document}